\documentclass{fairmeta}
\usepackage{amsmath}
\usepackage{enumerate} 
\usepackage{bm}
\usepackage{algorithm}
\usepackage{floatflt}
\usepackage{algpseudocode}
\usepackage{amsfonts}
\usepackage{amsthm}
\usepackage{newtxtt}
\usepackage{algorithm}
\usepackage{colortbl} 
\usepackage{cleveref}
\usepackage{diagbox} 
\usepackage[utf8]{inputenc}
\usepackage{textgreek}
\usepackage{colortbl}
\usepackage{nicematrix}
\usepackage{makecell}
\usepackage{float}
\usepackage{arydshln}
\usepackage[frozencache,cachedir=.]{minted}
\usepackage{caption}
\usepackage{subcaption}
\usepackage{tcolorbox}
\usepackage{amssymb}
\usepackage{xspace}
\usepackage{wrapfig}
\usepackage{adjustbox}
\usepackage{tabularx}
\usepackage{booktabs}
\usepackage{mathtools}
\usepackage{wrapfig}
\usepackage{amssymb}
\usepackage{graphicx}

\usepackage{silence}
\makeatletter
\patchcmd{\wrong@fontshape}{\@gobbletwo}{}{}{}
\makeatother
\definecolor{upColor}{RGB}{17,138,21}
\definecolor{downColor}{RGB}{174,36,67}

\newtheorem{theorem}{Theorem}[]

\newtheorem{remark1}[theorem]{Remark}

\title{The Law of Multi-Model Collaboration: Scaling Limits of Model Ensembling for Large Language Models}
\author[]{Dakuan Lu}
\author[]{Jiaqi Zhang}
\author[]{Cheng Yuan}
\author[]{Jiawei Shao}
\author[]{Xuelong Li}
\affiliation[]{Institute of Artificial Intelligence (TeleAI), China Telecom}
\metadata[Correspondence to]{Xuelong Li (\email{xuelong\_li@ieee.org})}

\begin{document}

\abstract{
Recent advances in large language models (LLMs) have been largely driven by scaling laws for individual models, which predict performance improvements as model parameters and data volume increase. However, the capabilities of any single LLM are inherently bounded. One solution originates from intricate interactions among multiple LLMs, rendering their collective performance surpasses that of any constituent model. Despite the rapid proliferation of multi-model integration techniques such as model routing and post-hoc ensembling, a unifying theoretical framework of performance scaling for multi-model collaboration remains absent.
In this work, we propose \textbf{the Law of Multi-model Collaboration}, a scaling law that predicts the performance limits of LLM ensembles based on their aggregated parameter budget. To quantify the intrinsic upper bound of multi-model collaboration, we adopt a method-agnostic formulation and assume an idealized integration oracle where the total cross-entropy loss of each sample is determined by the minimum loss of any model in the model pool.
Experimental results reveal that multi-model systems follow a power-law scaling with respect to the total parameter count, exhibiting a more significant improvement trend and a lower theoretical loss floor compared to single model scaling. Moreover, ensembles of heterogeneous model families achieve better performance scaling than those formed within a single model family, indicating that model diversity is a primary driver of collaboration gains. These findings suggest that model collaboration represents a critical axis for extending the intelligence frontier of LLMs.
}

\maketitle

\section{Introduction}

In recent years, large language models (LLMs) have made progress under the guidance of a unified theoretical framework: scaling laws that describe performance as a power-law function of model parameters, training data, and computation~\citep{kaplan2020scaling, henighan2020scaling}. These laws provide both forecasting power—letting practitioners predict performance before training begins—and a conceptual framework that treats intelligence progress as movement along a resource-constrained performance frontier. Refinements like compute-data trade-offs~\citep{hoffmann2022training} have cemented single-model scaling laws as foundational theory for LLM development. Yet these laws do more than summarize trends: they articulate boundaries of the single model. By characterizing how loss behaves asymptotically as scale increases, they implicitly define what a single model can achieve under ideal conditions~\citep{yang2025qwen3, grattafiori2024llama}. But this boundary has theoretical limitations. As models grow, performance gains show diminishing returns, demanding exponentially more resources for marginal improvements. Meanwhile, any single LLM is affected by its random initialization, architecture bias, and training data distribution, resulting in uneven performance across different tasks and domains, even among models of the same size~\citep{bender2021dangers, liang2022holistic}. This suggests that single-model scaling laws capture only one dimension along which intelligence can extend.

Motivated by these constraints, various multi-model integration approaches have been proposed~\citep{chen2025harnessing}—including dynamic routing~\citep{qian2025xroutertrainingcostawarellms}, ensemble voting~\citep{zhao2025majority}, tool orchestration~\citep{zhang2025router}, and agent-based collaboration~\citep{tran2025multi, chen2025tool, li2024survey, fan2025rethinking}—which demonstrate superior performance and efficiency over individual models across diverse application scenarios. 
But these successes remain empirical and method-specific, justified through heuristics or task evaluations rather than a comprehensive understanding of why multi-model collaboration works. 
The result is a serious mismatch between theory and practice: although the scaling law explains the growth behavior of individual models, there is no similar framework for multi-model collaboration. 
Without theoretical grounding, we cannot compare multi-model systems to large single models rigorously or reason about their ultimate limits. This gap motivates a general law for multi-model collaboration, parallel to existing single-model scaling laws~\citep{kaplan2020scaling, hoffmann2022training, kim2025pretraininginfinitecompute}.

We propose the Law of Multi-model Collaboration, which characterizes the performance limit of multi-model collaborative systems as a function of the total number of parameters of all models involved. Rather than focusing on specific ensemble mechanisms, we adopt a method-agnostic view, treating multi-model systems as collective entities. This study operates within the AI Flow framework~\citep{an2025ai, 10884554}, which advocates distributed intelligence integrating artificial intelligence with communication systems through collaborative interactions across heterogeneous models deployed at device, edge, and cloud levels~\citep{11222951}. Understanding fundamental performance limits of multi-model collaboration becomes a prerequisite for principled system design in this paradigm. Our formulation centers on a model pool of LLMs with total parameter count summed across the pool. For a given input distribution, we characterize the minimal achievable loss through model collaboration, defining a theoretical lower bound analogous to single-model scaling laws. To isolate intrinsic collaboration limits, we consider an idealized oracle integration where per-token cross-entropy loss equals the minimum achieved by any pool model. While unrealizable in practice, this oracle parallels optimality assumptions in prior scaling studies that abstract away algorithmic inefficiencies to reveal fundamental trends. We construct empirical performance frontiers by enumerating model combinations from a fixed pool, evaluating aggregated parameter budgets against achieved loss, and extracting Pareto-optimal points representing the theoretical performance envelope.

Our analysis yields two key findings. First, multi-model systems exhibit clear power-law scaling between loss and total parameter count, following a similar scaling exponent to single models ($\alpha \approx 0.35$), but achieving a dramatically lower asymptotic loss floor—a 43\% reduction from 2.21 to 1.25. This indicates that multi-model collaboration fundamentally shifts the achievable performance frontier downward, enabling access to performance regimes unattainable by any individual model regardless of scale. Second, scaling efficiency depends strongly on model diversity: pairwise analysis reveals that heterogeneous ensembles from different families~\citep{yang2025qwen3, grattafiori2024llama, team2025gemma} consistently outperform homogeneous same-family ensembles under equivalent parameter budgets, with cross-family combinations achieving substantially lower asymptotic loss floors. These findings imply model diversity constitutes a fundamental scaling dimension complementary to parameters, data, and compute. The Law of Multi-model Collaboration thus reframes ensemble methods from ad-hoc techniques into a systematic, theoretically grounded pathway for extending artificial intelligence frontiers. This work makes four contributions:
\begin{itemize}
    \item We propose the \emph{Law of Multi-model Collaboration}, the first scaling-law framework characterizing performance limits of LLM ensembles.
    \item We demonstrate that multi-model systems follow predictable power-law scaling with respect to total parameter budget, achieving significantly lower theoretical loss bounds than single-model scaling.
    \item We show model heterogeneity plays a decisive role in scaling efficiency, with diverse-family ensembles consistently outperforming homogeneous ensembles under equivalent parameter budgets.
    \item We establish principled theoretical comparison between single-model and multi-model performance frontiers, reframing ensemble methods as a systematic pathway for extending the intelligence scaling frontier.
\end{itemize}

\section{Problem Formulation and Theoretical Framework}

In this section, we formalize the problem setting and introduce the theoretical framework underlying the \emph{Law of Multi-model Collaboration}. Our goal is to characterize the fundamental performance limits of multi-model systems in a manner that is independent of specific integration algorithms.

\subsection{Preliminaries}

Let $\mathcal{D}$ denote a fixed text distribution over sequences $x = (x_1, \dots, x_T)$.
Let $\mathcal{M} = \{M_1, M_2, \dots, M_K\}$ be a pool of pretrained large language models.
Each model $M_i$ is associated with a parameter count $P_i$ and a tokenizer $\tau_i$, which maps an input text $x$ to a token sequence
\[
\tau_i(x) = (t^{(i)}_1, \dots, t^{(i)}_{n_i(x)}),
\]
where the resulting token sequence length $n_i(x)$ may vary across models due to tokenizer differences.

We denote by $p_{M_i}(t \mid \cdot)$ the next-token predictive distribution induced by model $M_i$ under its tokenizer.

\subsection{Text-level Loss under Heterogeneous Tokenization}

A key challenge in comparing different language models arises from the fact that distinct models employ different tokenization schemes, resulting in different token sequence lengths for the same input text.
To enable a fair comparison across models, we define the modeling loss at the \emph{text level}, rather than at the per-token average level.
Specifically, for a given model $M_i$ and input text $x$, we define the loss as the \emph{sum} of token-level cross-entropy losses:
\begin{equation}
\mathcal{L}(M_i, x)
\;=\;
\sum_{j=1}^{n_i(x)}
-\log p_{M_i}\!\left(t^{(i)}_j \,\middle|\, t^{(i)}_{<j}\right).
\label{eq:text_level_loss}
\end{equation}

To account for tokenization differences while maintaining comparability, we normalize the expected loss by the average token sequence length across all models and all test texts. Let $\bar{n}$ denote the average token sequence length over all models in the model pool and all texts in the evaluation dataset:
\begin{equation}
\bar{n} = \frac{1}{|\mathcal{M}| \cdot |\mathcal{D}|} \sum_{M_i \in \mathcal{M}} \sum_{x \in \mathcal{D}} n_i(x).
\end{equation}

The expected loss of model $M_i$ over the distribution $\mathcal{D}$ is then given by
\begin{equation}
\mathcal{L}(M_i)
\;=\;
\frac{1}{\bar{n}} \mathbb{E}_{x \sim \mathcal{D}}\left[ \mathcal{L}(M_i, x) \right].
\end{equation}

This normalization enables fair comparison across models with heterogeneous tokenizers while preserving the sensitivity to modeling efficiency at the text level.

\subsection{Oracle Multi-model Collaboration}

To isolate the intrinsic upper bound of multi-model collaboration, we consider an idealized \emph{oracle ensemble}.
This oracle abstracts away any practical limitations of model integration and serves solely as a theoretical construct for characterizing achievable performance limits.
For a given model pool $\mathcal{S} \subseteq \mathcal{M}$ and input text $x$, the oracle ensemble loss is defined as
\begin{equation}
\mathcal{L}_{\mathrm{oracle}}(\mathcal{S}, x)
\;=\;
\min_{M_i \in \mathcal{S}} \; \mathcal{L}(M_i, x).
\label{eq:oracle_loss}
\end{equation}

That is, for each individual text, the oracle selects the single model within the pool that yields the lowest text-level loss (sum of token-level losses) under its own tokenizer.
The expected oracle loss over the distribution $\mathcal{D}$ is then normalized by the average token sequence length $\bar{n}$:
\begin{equation}
\mathcal{L}_{\mathrm{oracle}}(\mathcal{S})
\;=\;
\frac{1}{\bar{n}} \mathbb{E}_{x \sim \mathcal{D}}
\left[
\mathcal{L}_{\mathrm{oracle}}(\mathcal{S}, x)
\right].
\end{equation}

We stress that the oracle ensemble does not correspond to any realizable inference-time system. Instead, it defines a \emph{best-case lower bound} on loss achievable by any multi-model collaboration mechanism operating over the model pool $\mathcal{S}$. This abstraction is directly analogous to idealized assumptions commonly employed in prior scaling-law analyses.

\subsection{Aggregated Parameter Budget}

For a model pool $\mathcal{S}$, we define its aggregated parameter budget as the sum of parameters across all constituent models:
\begin{equation}
P_{\mathrm{total}}(\mathcal{S})
\;=\;
\sum_{M_i \in \mathcal{S}} P_i.
\end{equation}

This quantity serves as the primary scaling variable for multi-model systems, paralleling the role of parameter count in single-model scaling laws.

\subsection{Multi-model Scaling Frontier}

Given a fixed model pool $\mathcal{M}$, we consider all possible subsets $\mathcal{S} \subseteq \mathcal{M}$ and evaluate their corresponding pairs
\[
\left(
P_{\mathrm{total}}(\mathcal{S}),
\;
\mathcal{L}_{\mathrm{oracle}}(\mathcal{S})
\right).
\]

Among these points, we identify the \emph{Pareto-optimal frontier}, consisting of all subsets for which no other subset achieves both lower loss and lower total parameter count.
This frontier represents the theoretical performance envelope of multi-model collaboration under the oracle assumption.
The object of study in this work is the functional relationship between
$P_{\mathrm{total}}(\mathcal{S})$ and the minimal achievable loss on this frontier.

\subsection{Law of Multi-model Collaboration}

We are now ready to state the \emph{Law of Multi-model Collaboration}.
Empirically, we observe that the Pareto-optimal oracle loss follows a power-law relationship with respect to the aggregated parameter budget:
\begin{equation}
\mathcal{L}_{\mathrm{oracle}}(P)
\;\approx\;
A \, P^{-\alpha} + \mathcal{L}_{\infty},
\label{eq:multimodel_scaling_law}
\end{equation}
where $A > 0$ is a scaling constant, $\alpha > 0$ is the scaling exponent, and $\mathcal{L}_{\infty}$ denotes the asymptotic loss floor.

This law generalizes classical single-model scaling laws to the multi-model setting. Crucially, both the exponent $\alpha$ and the loss floor $\mathcal{L}_{\infty}$ depend on the composition of the model pool, with the trade-off between scaling rate and asymptotic performance being fundamentally shaped by model diversity. While homogeneous pools may exhibit steeper scaling exponents at smaller budgets, heterogeneous pools achieve substantially lower asymptotic loss floors, reflecting complementary modeling capabilities across architectural families.

\section{Experimental Protocol for Multi-model Scaling}

In this section, we describe the experimental protocol used to empirically characterize the scaling behavior of multi-model collaboration. The design of this protocol closely follows the methodology established in prior single-model scaling law studies, while extending it to the multi-model setting defined in Section~2.

\subsection{Model Pool Construction}

We construct a model pool $\mathcal{M}$ consisting of $71$ publicly available open-source \emph{base} language models.
The pool spans a wide range of parameter scales and includes multiple widely adopted model families, such as \textsc{Qwen}, \textsc{LLaMA}, \textsc{Gemma}, among others.
All models are evaluated in their base (pre-alignment) form to minimize confounding effects introduced by instruction tuning or reinforcement learning from human feedback.
The selected models exhibit substantial diversity in architecture, training data composition, and optimization strategies, resulting in a broad spectrum of modeling capabilities.
This diversity enables a systematic investigation of both homogeneous and heterogeneous multi-model collaboration.

\subsection{Evaluation Dataset and Loss Measurement}

All models are evaluated on a fixed held-out text dataset sampled from the distribution $\mathcal{D}$.
For each model $M_i \in \mathcal{M}$ and each input text $x$, we compute the text-level loss $\mathcal{L}(M_i, x)$ as defined in Equation~\eqref{eq:text_level_loss}, namely the sum of token-level cross-entropy losses under the model's own tokenizer.
To obtain the expected loss $\mathcal{L}(M_i)$, we first compute the average of the summed token losses over all texts in the evaluation dataset, then normalize by the average token sequence length $\bar{n}$ across all models and all test texts.
This normalization procedure yields a single-model performance baseline that is directly comparable across models despite heterogeneous tokenization schemes, while accounting for differences in tokenizer granularity.

\subsection{Enumeration of Multi-model Combinations}

To explore the space of multi-model collaboration while addressing the combinatorial explosion of possible subsets, we adopt a Pareto-pruning strategy.
Directly enumerating all $2^{71} - 1$ non-empty subsets of the model pool is computationally prohibitive.
Instead, we employ an iterative procedure that constructs ensembles of size $k$ by incrementally adding models to Pareto-optimal ensembles of size $k-1$.

Specifically, we proceed as follows:
\begin{enumerate}
    \item For $k=1$, evaluate all single models and extract the Pareto-optimal frontier based on parameter count and oracle loss.
    \item For $k \geq 2$, consider only ensembles formed by adding one model from $\mathcal{M}$ to a Pareto-optimal ensemble of size $k-1$.
    \item For each candidate ensemble $\mathcal{S}$, compute the aggregated parameter budget $P_{\mathrm{total}}(\mathcal{S})$ and the oracle loss $\mathcal{L}_{\mathrm{oracle}}(\mathcal{S})$ as defined in Equation~\eqref{eq:oracle_loss}.
    \item Extract the Pareto-optimal frontier among all ensembles of size $k$, and proceed to $k+1$.
\end{enumerate}

This iterative Pareto-pruning procedure enables us to explore ensembles ranging from 2 to 71 models while maintaining computational feasibility.
The resulting configurations represent a representative sample of the multi-model scaling frontier, capturing the fundamental scaling trends without exhaustive enumeration.

To further investigate the role of model diversity, we also analyze pairwise combinations ($k=2$) separately, categorizing them into:
\begin{enumerate}
    \item \textbf{Homogeneous pairs}, where both models belong to the same model family;
    \item \textbf{Heterogeneous pairs}, where the models originate from different model families.
\end{enumerate}

This partition enables a controlled comparison of scaling efficiency with and without architectural and training diversity at the pairwise level.

\subsection{Pareto-optimal Frontier Extraction}

Due to the large number of models and the wide variance in their capabilities, the raw set of evaluated points contains many dominated configurations.
A configuration $\mathcal{S}$ is considered Pareto-optimal if no other configuration achieves both a lower oracle loss and a lower aggregated parameter budget.

To characterize the fundamental scaling behavior of multi-model collaboration, we extract Pareto-optimal frontiers for:
\begin{itemize}
    \item \textbf{Single-model frontier}: consisting of 12 Pareto-optimal models selected from the 71 base models;
    \item \textbf{Multi-model ensemble frontier}: consisting of 3,146 Pareto-optimal ensembles selected from all explored combinations ranging from 2 to 71 models, obtained via the iterative Pareto-pruning procedure described above.
\end{itemize}

Additionally, to investigate the impact of model diversity on scaling efficiency, we separately extract Pareto-optimal frontiers from pairwise combinations:
\begin{itemize}
    \item \textbf{Homogeneous pairs}: 22 Pareto-optimal pairs selected from 506 same-family combinations;
    \item \textbf{Heterogeneous pairs}: 32 Pareto-optimal pairs selected from 1,979 cross-family combinations.
\end{itemize}

The resulting Pareto frontiers represent the empirical lower envelopes of achievable loss under the oracle collaboration assumption.

\begin{figure}[htb]
    \centering
    \includegraphics[width=\linewidth]{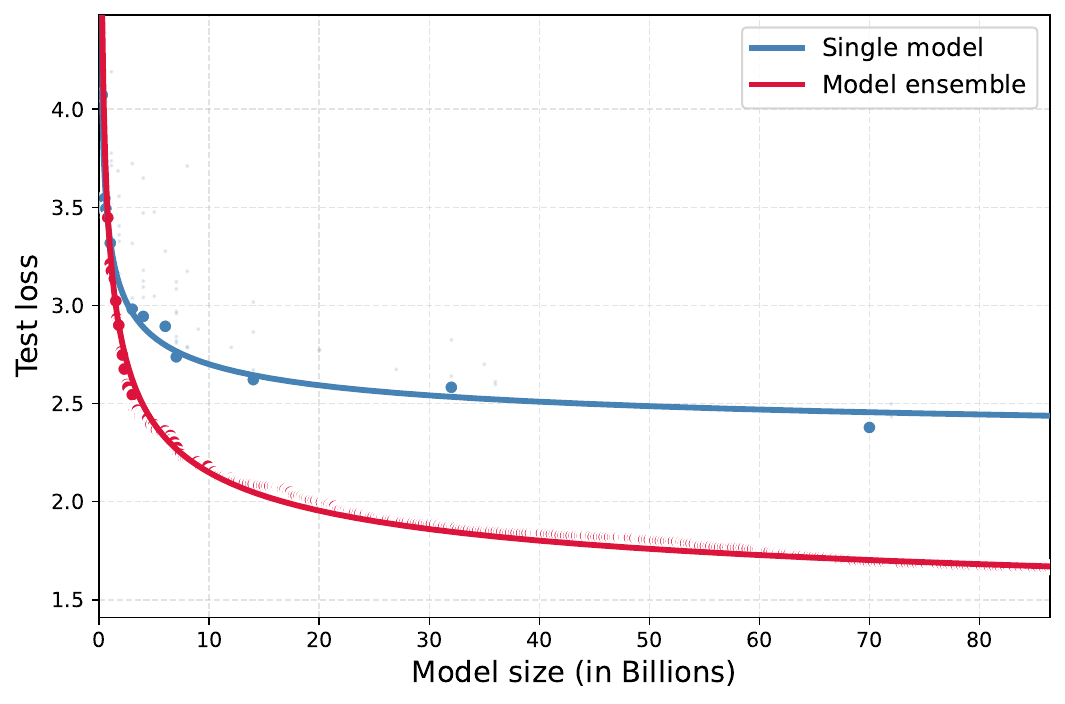}
    \caption{Comparison between scaling performance under two settings: single model and multi-model ensemble. 
    Solid curves denote nonlinear least-squares fits of the scaling formula, while points correspond to Pareto-optimal lower envelopes.
    Model ensembling requires fewer parameters to achieve a target loss and reduces the lower bound of the test loss significantly.}
    \label{fig:scaling_law_envelope}
\end{figure}

\subsection{Scaling Law Fitting}

For each Pareto-optimal frontier, we fit the scaling law of the form
\begin{equation}
\mathcal{L}(P)
=
A \, P^{-\alpha} + \mathcal{L}_{\infty},
\end{equation}
as introduced in Equation~\eqref{eq:multimodel_scaling_law}.
The parameters $(A, \alpha, \mathcal{L}_{\infty})$ are estimated using nonlinear least squares regression on the Pareto-optimal points.
Fitting is performed in the original loss domain rather than the log-transformed domain to ensure consistency with prior scaling-law analyses.
This procedure yields comparable scaling exponents and asymptotic loss estimates for single-model, homogeneous multi-model, and heterogeneous multi-model settings, enabling a direct empirical comparison of their scaling behaviors.

\section{Empirical Scaling Laws of Multi-model Collaboration}

In this section, we present empirical evidence supporting the \emph{Law of Multi-model Collaboration}.
By analyzing the Pareto-optimal performance frontiers constructed in Section~3, we demonstrate that multi-model systems exhibit stable and predictable power-law scaling behavior with respect to aggregated parameter budgets, and that this behavior differs systematically from that of single models.

\subsection{Scaling Behavior of Single and Multi-model Systems}

Figure~\ref{fig:scaling_law_envelope} compares the Pareto-optimal lower envelopes for single models and multi-model ensembles.
The single-model frontier consists of 12 Pareto-optimal configurations selected from the 71 base models, while the multi-model ensemble frontier comprises 3,146 Pareto-optimal configurations obtained through the iterative Pareto-pruning procedure over ensemble sizes ranging from 2 to 71 models.
Each point on the envelope corresponds to a Pareto-optimal configuration in terms of aggregated parameter count and oracle loss, while the solid curves represent nonlinear least-squares fits of the scaling law defined in Equation~\eqref{eq:multimodel_scaling_law}.
Both settings exhibit a clear power-law relationship between loss and total parameter count, demonstrating that multi-model systems exhibit regular and predictable scaling behavior analogous to classical single-model scaling laws, despite their increased structural complexity.

\begin{table}[t]
\centering
\caption{Fitted scaling law parameters for single-model and multi-model ensemble frontiers. The loss floor $L_{\infty}$ is significantly reduced for model ensembles, signifying the benefit of multi-model collaboration. The multi-model ensemble frontier aggregates all explored combinations from 2 to 71 models.}
\label{tab:scaling_law_results}
\begin{tabular}{lccccc}
\toprule
\textbf{Setting} & \textbf{Raw Points} & \textbf{Pareto Points} & $L_{\infty}$ & $A$ & $\alpha$ \\
\midrule
Single model & 71 & 12 & 2.21 & 1.11 & 0.3578 \\
Model ensemble & 3,146 & 3,146 & 1.25 & 2.02 & 0.3502 \\
\bottomrule
\end{tabular}
\end{table}

\subsection{Quantitative Comparison of Scaling Exponents}

To quantify differences in scaling efficiency, we fit the scaling law to the Pareto-optimal frontiers for both single-model and multi-model ensemble settings.
Table~\ref{tab:scaling_law_results} reports the fitted parameters, including the scaling exponent $\alpha$ and the asymptotic loss floor $L_{\infty}$.
Single-model scaling yields an exponent of $\alpha = 0.3578$, consistent with previously reported scaling behavior in language modeling.
The multi-model ensemble frontier, which aggregates all explored combinations from 2 to 71 models, exhibits a comparable scaling exponent of $\alpha = 0.3502$.
While the scaling rates are similar, the critical distinction lies in the asymptotic behavior.
The multi-model ensemble achieves a substantially lower loss floor of $L_{\infty} = 1.25$, compared to $L_{\infty} = 2.21$ for single models—a reduction of approximately 43\%.
This pronounced decrease indicates that multi-model collaboration fundamentally shifts the achievable performance frontier downward, enabling access to regions of the loss landscape unattainable by any individual model.

Specifically, the fitted scaling laws are given by:
\begin{align}
\mathcal{L}_{\text{single}}(P) &= \frac{1.11}{P^{0.3578}} + 2.21, \label{eq:single_model_fitted} \\
\mathcal{L}_{\text{ensemble}}(P) &= \frac{2.02}{P^{0.3502}} + 1.25, \label{eq:ensemble_fitted}
\end{align}
where $P$ denotes the total parameter count (in billions).

\subsection{Asymptotic Loss Floors}

The fitted asymptotic loss floors $L_{\infty}$ reveal the fundamental difference between single-model and multi-model scaling regimes.
Single-model scaling exhibits an estimated loss floor of $L_{\infty} = 2.21$, representing the theoretical lower bound achievable by any individual model architecture under the oracle assumption.
In contrast, the multi-model ensemble frontier achieves a substantially lower asymptotic loss floor of $L_{\infty} = 1.25$, representing a 43\% reduction.
This pronounced reduction indicates that multi-model collaboration enables access to complementary modeling capabilities that are unattainable by scaling a single architecture alone.
The lower loss floor suggests that different models encode distinct hypotheses about the data distribution, and that their oracle combination effectively expands the representational capacity of the system.
As the aggregated parameter budget increases, the multi-model ensemble asymptotically approaches a performance regime fundamentally inaccessible to any single model, regardless of its size.

\subsection{The Role of Model Diversity: Homogeneous versus Heterogeneous Pairs}

\begin{figure}[htb]
    \centering
    \includegraphics[width=\linewidth]{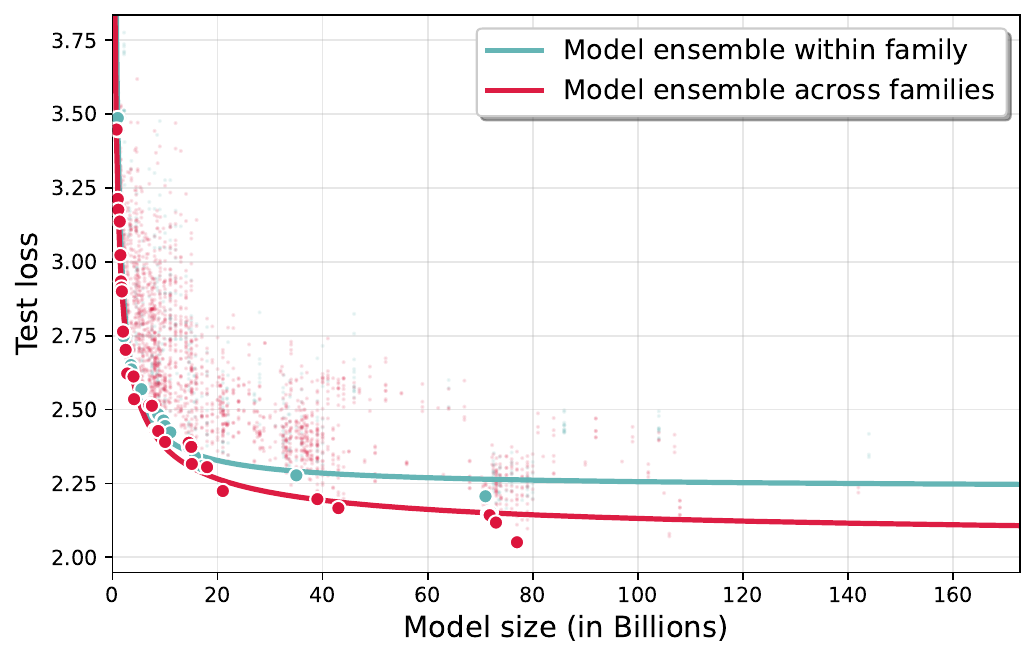}
    \caption{Comparison between scaling performance between model ensemble within the same family and that across different families. 
    Solid curves denote nonlinear least-squares fits of the scaling formula, while points correspond to Pareto-optimal lower envelopes.
    Collaboration beyond a single model series results in better scaling performance, validating the complementarity among different model series.}
    \label{fig:scaling_law_same_and_diff}
\end{figure}

To investigate the impact of model diversity on collaboration efficiency, we separately analyze pairwise combinations, as illustrated in Figure~\ref{fig:scaling_law_same_and_diff}.
Among the 506 same-family pairs, 22 are Pareto-optimal; among the 1,979 cross-family pairs, 32 are Pareto-optimal.

The Pareto frontiers reveal distinct scaling behaviors, as summarized in Table~\ref{tab:pairwise_scaling}.
Homogeneous pairs (same-family combinations) exhibit a steep scaling exponent ($\alpha = 0.8501$), reflecting rapid initial improvements at small parameter budgets.
However, their performance quickly saturates at an asymptotic loss floor of $L_{\infty} = 2.2314$, which is comparable to that of single models ($L_{\infty} = 2.21$).
This behavior suggests that combining architecturally similar models primarily reduces variance or redundancy within a shared inductive bias, rather than fundamentally extending representational capacity.

In contrast, heterogeneous pairs (cross-family combinations) exhibit a more moderate scaling exponent ($\alpha = 0.5516$) but achieve a substantially lower asymptotic loss floor of $L_{\infty} = 2.0377$.
This indicates that architectural and training diversity is a primary driver of collaboration gains.
While same-family pairs achieve rapid short-term improvements through redundancy reduction, cross-family pairs unlock complementary modeling capabilities that lower the fundamental performance limit.
Models from different families encode distinct hypotheses about the data distribution, and their combination enables access to a broader region of the hypothesis space.

These findings demonstrate that model diversity is not merely a practical consideration but a fundamental factor governing the efficiency of multi-model scaling.
The trade-off between steeper short-term scaling (homogeneous) and lower asymptotic limits (heterogeneous) highlights the importance of diversity in achieving sustained performance gains.

\begin{table}[t]
\centering
\caption{Fitted scaling law parameters for pairwise model combinations. Same-family pairs exhibit steeper scaling exponents but saturate at loss floors comparable to single models, while cross-family pairs achieve substantially lower asymptotic loss through architectural diversity.}
\label{tab:pairwise_scaling}
\begin{tabular}{lccccc}
\toprule
\textbf{Combination Type} & \textbf{Raw Pairs} & \textbf{Pareto Pairs} & $L_{\infty}$ & $A$ & $\alpha$ \\
\midrule
Same-family pairs & 506 & 22 & 2.2314 & 1.2281 & 0.8501 \\
Cross-family pairs & 1,979 & 32 & 2.0377 & 1.1885 & 0.5516 \\
\bottomrule
\end{tabular}
\end{table}

\subsection{Evidence for the Law of Multi-model Collaboration}

Taken together, these results constitute empirical evidence for the \emph{Law of Multi-model Collaboration}.
Multi-model systems exhibit predictable power-law scaling with respect to aggregated parameter budgets, while achieving systematically improved scaling efficiency and lower theoretical loss bounds compared to single-model scaling.
Crucially, the composition of the model pool plays a central role in determining scaling behavior.
Diversity across model families emerges as a key factor in extending the attainable performance frontier, highlighting multi-model collaboration as a complementary and orthogonal scaling axis to parameters, data, and compute.

\section{Interpretation and Theoretical Implications}

In this section, we interpret the empirical results given in Section~4 and discuss their broader theoretical implications.
Instead of proposing new algorithms, we want to know what the observed scaling behavior can tell us about how different models work together and if it matches up with any ideas people already have about how models should behave when they get bigger.

\subsection{Multi-model Collaboration as an Extension of Scaling Laws}

Classical scaling laws characterize how the performance of a \emph{single} model improves as a function of its parameter count, data, and compute.
Implicit in these laws is the assumption that scaling occurs along a single architectural and representational trajectory~\citep{kaplan2020scaling, hoffmann2022training}.
Our results indicate that multi-model collaboration is a qualitatively distinct kind of scaling.
There is a stable power law relationship between oracle loss and aggregated parameter budget for a multi-model system, which means that the multi-model system has its own scaling regime.
This regime is not just a simple pile-up of single-model behavior.
On the contrary, better scaling exponents and lower asymptotic loss floors show that working together lets us reach parts of the performance frontier that we can’t get to just by making one model bigger.
And so this leads us to think of the total parameter budget of a pool of models as a new kind of scaling variable that’s somewhat similar to the number of parameters for a single model, but it’s living in a much bigger model space.

\subsection{Interpretation of Scaling Exponents}

The scaling exponent $\alpha$ reflects the efficiency with which additional parameters translate into performance gains.
Notably, the scaling exponents for single-model and multi-model ensemble frontiers are remarkably similar ($\alpha = 0.3578$ and $\alpha = 0.3502$, respectively), indicating that the \emph{rate} of loss reduction with respect to parameter count follows a consistent power-law across both regimes.
This similarity suggests that the fundamental efficiency of parameter utilization, in terms of marginal gains per additional parameter, is comparable whether parameters are concentrated within a single architecture or distributed across multiple models.
However, the critical distinction lies not in the scaling rate, but in the \emph{baseline performance level} from which scaling occurs.
Multi-model collaboration effectively shifts the entire performance curve downward, achieving lower loss at every parameter budget.
This vertical shift manifests most dramatically in the asymptotic regime, where the multi-model ensemble approaches a substantially lower loss floor than any single model.
In the pairwise diversity analysis, we observe a revealing trade-off between scaling rate and asymptotic performance.
Homogeneous same-family pairs exhibit a steep scaling exponent ($\alpha = 0.8501$), reflecting rapid early gains at small parameter budgets, but quickly saturate at an asymptotic loss floor ($L_{\infty} = 2.2314$) comparable to single models ($L_{\infty} = 2.21$).
This suggests that combining architecturally similar models primarily reduces variance or redundancy within a shared inductive bias without fundamentally expanding representational capacity.
In contrast, heterogeneous cross-family pairs exhibit a more moderate scaling exponent ($\alpha = 0.5516$) but achieve a substantially lower asymptotic loss floor ($L_{\infty} = 2.0377$).
This pattern is consistent with the notion that different model families encode complementary hypotheses about the data distribution~\citep{dietterich2000ensemble}.
The steeper exponent of homogeneous pairs indicates efficient exploitation of redundancy, while the lower floor of heterogeneous pairs reveals the value of architectural diversity in accessing distinct regions of the hypothesis space.

\subsection{Lower Loss Floors and the Limits of Single-Model Scaling}

The 43\% reduction in asymptotic loss floor observed for multi-model ensembles (from 2.21 to 1.25) has profound implications for our understanding of the limits of single-model scaling.
In classical scaling laws, the loss floor is often interpreted as reflecting irreducible error under a fixed model class and training regime~\citep{kaplan2020scaling, henighan2020scaling}.
Our findings demonstrate that this floor is not an absolute boundary but rather a constraint specific to individual architectural lineages.
The dramatic reduction in loss floor reveals that multi-model collaboration fundamentally expands the hypothesis space accessible to the system.
While a single model, regardless of its size, is constrained by the inductive biases inherent in its architecture and training procedure, an oracle ensemble can adaptively leverage complementary strengths across multiple architectures.
This suggests that the variety of inductive biases available to the system not merely the total number of parameters determines the achievable performance limit.
From this perspective, the loss floor for single-model scaling laws should be understood as architecture-specific rather than representing a fundamental limit of the modeling task itself.
Multi-model collaboration effectively transcends this constraint by accessing a broader region of the hypothesis space~\citep{domingos2012few}, enabling performance levels that are structurally inaccessible to any individual architecture.

\subsection{Relation to Existing Paradigms}

Multi-model collaboration shares superficial similarities with ensemble learning and mixture-of-experts methods, but differs in important respects.
Traditional ensembles~\citep{chen2025harnessing} are typically designed to reduce variance or improve robustness around a fixed model class, while mixture-of-experts~\citep{10937907} architectures are trained jointly with explicit routing mechanisms.
In contrast, the framework studied here considers pretrained models as fixed entities and focuses on the theoretical limits of their collective performance under idealized coordination.
The oracle formulation removes the training dynamics and routing mechanism from consideration, so that the scaling behavior itself can be studied on its own.
Therefore, the Law of Multi-model Collaboration can be regarded as supplementary to, but not a substitute for, existing single-model scaling theories.

\subsection{Implications for the Intelligence Frontier}

Taken together, these results suggest that progress in language modeling need not rely exclusively on scaling individual models along a single dimension.
Model diversity becomes a new orthogonal direction that the intelligence frontier can expand into.
The Law of Multi-Model Collaboration transforms multi-model systems from heuristic engineering constructions into objects that follow consistent, predictable scaling rules.
It means that we could make progress in the future without necessarily having to train bigger models, but rather by cleverly combining different kinds of models so they can work well together.

\section{Limitations and Scope of the Law}

While the empirical results presented in this work provide strong evidence for the \emph{Law of Multi-model Collaboration}, it is important to clarify the scope of the law and acknowledge its limitations.
These limitations arise from the theoretical perspective taken here, and they do not detract from the main contribution, which is identifying the basic performance limits rather than suggesting practical systems.

\subsection{Idealized Oracle Assumption}

An important abstraction in our formulation is the oracle ensemble, which selects, for each input text, the model within a pool that achieves the lowest loss.
It is an intentionally idealized assumption that does not correspond to any feasible inference time mechanism.
In practice, it is usually impossible to determine which model performs best for each input without knowing the ground-truth loss.
However, this abstraction serves a role analogous to idealized assumptions commonly employed in prior scaling-law analyses, such as optimal data usage or perfect optimization.
The oracle ensemble sets an upper limit on the performance that can be achieved with multi-model collaboration, thus enabling the study of intrinsic scaling behavior without considering algorithmic inefficiency.
Therefore, our results should be seen as theoretical limits rather than deployable architectures.

\subsection{Upper Bounds versus Realizable Systems}

The Law of Multi-model Collaboration describes the behavior of Pareto-optimal lower envelopes under an oracle assumption.
As such, it does not claim that practical multi-model systems can attain the same performance.
Real-world implementations have other constraints, such as routing errors, coordination overhead, latency, and compute costs.
Nonetheless, upper-bound analyses play a critical role in guiding system design by revealing what is possible in principle.
Like how single-model scaling laws give us ideas on what kind of models to use and where to put the data, even though they make some idealized assumptions, this law gives us something to compare against and encourages people to work on better ways of putting different models together.

\subsection{Pareto-Pruning as an Approximation Strategy}

To address the combinatorial explosion inherent in exploring all $2^{71} - 1$ possible model subsets, we employ an iterative Pareto-pruning strategy that constructs size-$k$ ensembles by incrementally adding models to Pareto-optimal ensembles of size $k-1$.
While this greedy heuristic enables tractable exploration of the multi-model scaling frontier, it does not guarantee exhaustive coverage of all potentially optimal configurations.
Specifically, it is possible that some non-Pareto-optimal ensembles of size $k-1$ could, when augmented with appropriate models, yield Pareto-optimal ensembles of size $k$ that are missed by our procedure.
Nonetheless, the resulting 3,146 Pareto-optimal configurations exhibit stable and predictable power-law scaling behavior, suggesting that the pruning strategy captures the fundamental scaling trends.
The close agreement between the empirical frontier and the fitted scaling law provides evidence that the explored configurations are representative of the broader multi-model scaling regime.
Future work could investigate alternative search strategies, such as genetic algorithms or reinforcement learning-based exploration, to more comprehensively sample the multi-model configuration space.

\subsection{Dependence on the Model Pool}

The fitted scaling parameters rely on the makeup of the underlying model pool.
Different choices for model families, parameter ranges, or training protocols can cause differences in scaling exponents and loss floors.
Accordingly, the numerical values reported in this work should not be interpreted as universal constants.
Rather, the law asserts the existence of a stable power-law relationship and systematic differences between single-model and multi-model scaling regimes.
The qualitative conclusions about enhanced scaling efficiency and decreased asymptotic loss with heterogeneous collaboration hold up well for all kinds of models.

\subsection{Loss Definition and Tokenization Effects}

To enable fair comparison across models with heterogeneous tokenizers, we compute the text-level loss as the sum of token-level cross-entropy values, which is then normalized by the average token sequence length across all models and all test texts.
This formulation preserves sensitivity to modeling efficiency at the text level while accounting for tokenization granularity differences.
The normalization by a shared average sequence length ensures that models are evaluated on a consistent scale, avoiding biases introduced by tokenizer-specific sequence lengths.
While this formulation is well-suited for comparing oracle performance bounds, alternative loss normalizations may be more appropriate in other contexts.
The law itself is independent of the particular choice of the loss function, and it will be interesting for future work to study the effect of different evaluation metrics on the observed scaling behavior.

\subsection{Scope of Applicability}

Lastly, we stress that the Law of Multi-model Collaboration is meant to supplement, not supplant, current single-model scaling theories. Single model scaling still matters for optimization, training dynamics, and resource allocation inside one fixed architecture. The law proposed here operates at a different level of abstraction, focusing on how multiple pretrained models interact as collective entities.
Within this scope, the law provides a principled framework for reasoning about multi-model systems and highlights collaboration and diversity as fundamental dimensions of scaling.

\section{Conclusion}


We introduce the \emph{Law of Multi-model Collaboration}, a scaling law defining the theoretical performance limits of systems composed of multiple pretrained LLMs. By employing an oracle-based, method-agnostic formulation, we isolate the intrinsic behavior of multi-model collaboration and provide a principled comparison to classical single-model scaling laws.

Through systematic exploration of model combinations ranging from 2 to 71 models via iterative Pareto-pruning, we demonstrate that multi-model systems follow stable power-law scaling with respect to aggregated parameter budgets.
While exhibiting similar scaling exponents to single models ($\alpha \approx 0.35$), multi-model ensembles achieve a dramatic 43\% reduction in asymptotic loss floor (from 2.21 to 1.25), fundamentally shifting the achievable performance frontier downward.
This indicates that multi-model collaboration does not merely accelerate convergence along the single-model trajectory but rather enables access to performance regimes that are structurally inaccessible to any individual architecture.

Additionally, pairwise analysis reveals that model diversity is a critical driver of collaboration gains: heterogeneous cross-family ensembles consistently outperform homogeneous same-family ensembles under equivalent parameter budgets, achieving substantially lower asymptotic loss floors.
These results suggest that distributing parameters across multiple models—especially those with diverse architectural and training characteristics—can be more effective than monolithic scaling within a single architecture.
Beyond the empirical results, this work reframes multi-model systems as objects governed by regular and predictable scaling behavior rather than as ad-hoc engineering solutions. 
The proposed law identifies model diversity as a fundamental scaling dimension, alongside parameters, data, and compute. From this perspective, the perceived limits of single-model scaling appear to be conditional on architectural lineage rather than absolute boundaries.
While the Law of Multi-model Collaboration defines theoretical upper bounds rather than realizable systems, such bounds are essential for clarifying what is achievable in principle. We hope this work inspires further research into diversity-aware scaling, principled integration mechanisms, and broader theories of collective intelligence in machine learning.

\bibliographystyle{plainnat}
\bibliography{paper}

\end{document}